\documentclass[runningheads]{llncs}
\usepackage[T1]{fontenc}
\usepackage{graphicx}
\usepackage{todonotes}
\usepackage{csquotes}
\usepackage{booktabs}
\usepackage{multirow}
\usepackage{makecell}
\usepackage{multirow}
\usepackage{longtable}
\usepackage{graphicx}
\usepackage{tabularx}

\usepackage{listings}
\usepackage{xcolor}
\begin{document}

\title{Better Call Claude: Can LLMs Detect Changes of Writing Style?}

\author{Johannes Römisch\inst{1} \and
Svetlana Gorovaia \inst{3} \and
Mariia Halchynska\inst{1} \and Gleb Schmidt\inst{2} \and Ivan P. Yamshchikov \inst{1}}

\authorrunning{Römisch et al.}

\institute{Center for Artificial Intelligence, Technical University of Applied Sciences Würzburg-Schweinfurt,  Münzstraße 12, 97070, Würzburg, Germany \and
Humanities Lab, Faculaty of Arts, Radboud University, Houtlaan 4, 6525 XZ, Nijmegen, Netherlands \and
LEYA Lab, School of Computer Science, Physics and Technology, HSE University, 6, 25th Liniya, Vasilievsky Ostrov, 199004, St Petersburg, Russia}

\maketitle            

\begin{abstract}
This article explores the zero-shot performance of state-of-the-art large language models (LLMs) on one of the most challenging tasks in authorship analysis: sentence-level style change detection. Benchmarking four LLMs on the official PAN~2024 and 2025 \textquote{Multi-Author Writing Style Analysis} datasets, we present several observations. First, state-of-the-art generative models are sensitive to variations in writing style---even at the granular level of individual sentences. Second, their accuracy establishes a challenging baseline for the task, outperforming suggested baselines of the PAN competition. 
Finally, we explore the influence of semantics on model predictions and present evidence suggesting that the latest generation of LLMs may be more sensitive to content-independent and purely stylistic signals than previously reported.

\end{abstract}

\section{Introduction}
Style change detection is one of the most challenging problems within the broader field of authorship analysis, with numerous academic and industrial applications ranging from philological and historical research to anti-plagiarism, copyright protection, forensics, cybersecurity, and governance. It is hardly a coincidence, therefore, that specific instances of this problem---such as author diarization or multi-author style analysis---have been featured among the PAN shared tasks since 2016, longer than any other task.

Approaches to this task primarily relied on feature engineering and classical machine learning and, more recently, deep learning methods, especially pre-trained model based solutions. Despite the ever-increasing quality and performance of these models, developing and using such systems has become prohibitively expensive, requiring large amounts of rare labeled data and advanced technical expertise. At the same time, these systems are often difficult to interpret---an important limitation for many use cases, particularly in academic research within the humanities and social sciences, as well as in forensic applications.

The release of a pleiade of LLMs since $2022$ is believed to have shifted the paradigm. Trained on extensive corpora, these models can detect even subtle regularities in texts, resulting in impressive performance in text generation and a variety of other downstream tasks. 

In the field of authorship analysis too, LLMs gave birth to entirely new fields of analysis, and the wave of enthusiasm among scholars suggests that LLMs are expected to soon bridge  the efficiency of deep learning systems and the need for interpretability.

Aiming to contribute to a better understanding of LLMs as tools for authorship analysis and to encourage their adoption, we present an analysis of their performance on sentence-level style change detection---a subfield of authorship analysis in which the capabilities of LLMs have so far remained underexplored. 

\subsection{Task Definition}

The core task of style change detection is to determine whether a given text was written by multiple authors and, if so, to identify the positions at which authorship changes occur\footnote{The task can be formulated at different levels of granularity (sentence, paragraph, etc.), but can also be presented as a clustering  problem.}. 

\subsection{Goals}
The first goal of this article is to survey the zero-shot performance of state-of-the-art generative AI models on the task. Our second objective is to explore various factors that influence the models’ predictions. We use the Hamming distance as a measure of response correctness and examine its correlation with parameters of the problems (e.g., number of authors, number of changes, length of the problem) and their semantic features. 

\section{Related Work}
\subsection{Style Change Detection}
In early editions of PAN, the task was approached through manual and corpus-specific engineering of stylistic features subsequently used for unsupervised clustering of predefined text segments \cite{stamatatos_clustering_2016,potthast_overview_2017,gomez-adorno_author_2017,kocher_author_2017}. Since the late 2010s, methodologies shifted towards neural networks \cite{bagnall_authorship_2016}, and by 2023, transformer-based architectures dominated approaches presented at PAN. 

Leveraging rich linguistic knowledge from pre-training on massive corpora, transformer backbones consistently achieve exceptional performance, regularly surpassing $80$\% accuracy even on challenging datasets with uniform topic signal. Apart from traditional classification \cite{hashemi_enhancing_2023,lin_team_2024}, particularly popular strategies include contrastive learning \cite{zangerle_overview_2023,ye_supervised_2023,wu3-2024}, ensembling of models fine-tuned on various aspects of style. 

Less conventional yet insightful methodologies also were presented by employing manually-constructed prompts with masked language modeling \cite{zhang_style_2022}; Gao et al. \cite{graner_unorthodox_2022} openly \textquote{hacked} the task using extrinsic web clues. 

Additionally, closely related tasks such as detecting AI-generated or Human-AI co-authored texts share these methodologies as well as boundary detection and segment-level attribution. For instance, \cite{zheng-2024-detecting-ai} introduced a modular framework combining segment detection with sentence-level classification using various Transformer architectures \cite{zeng2024detectingaigeneratedsentenceshumanai,lo-etal-2021-transformer-pre,he2021deberta,zhuang-etal-2021-robustly}, while GigaCheck \cite{tolstykh2024gigacheckdetectingllmgeneratedcontent} combined sentence-level binary classification with DETR-style character-level span prediction for detailed LLM attribution. Other recent approaches have exploited structural and attributional signals beyond superficial content, such as TopFormer’s integration of topological features via Topological Data Analysis (TDA) \cite{uchendu-2024}, persistent homology derived from attention maps \cite{kushnareva-etal-2021-artificial}, and token-level log-probability dynamics aggregated through CNN and self-attention mechanisms \cite{wang-etal-2023-seqxgpt}. These methods, especially those focused on stylometric shifts, segmentation, or local attribution, are readily adaptable to style change detection.

\subsection{AI-assisted Authorship Analysis}
The surge of GenAI gave birth to a new field within authorship analysis---AI-assisted authorship analysis. New LLM-based methodologies emerged using LLMs for feature extraction, data augmentation, and direct stylistic analysis \cite{huang_authorship_2025}.

Relying on the robust abilities of LLMs to describe writing styles, \cite{patel_learning_2023} introduced a system producing interpretable style embeddings. Similarly, \cite{ramnath_cave_2024} used GPT-4-Turbo to generate structured stylistic descriptions as training data for a smaller Llama-3-8B model, enhancing interpretability. 

Directly prompting LLMs with authorship-related queries, \cite{huang_can_2024} and \cite{schmidt_sui_2024} report promising reliability in authorship verification, although the latter highlights potential bias due to semantic similarity in historical languages. Explicitly guiding models with linguistically-informed prompts (LIP) notably boosts performance and analytical quality, particularly in English texts \cite{huang_can_2024}. The PromptAV framework similarly directs LLMs through stylometric reasoning for authorship tasks \cite{hung-etal-2023-wrote}. 

Our experiment thus aligns with broader explorations of LLM capabilities in authorship analysis, contributing specifically to the niche but increasingly relevant task of sentence-level style change detection. 

\section{Methodology}
\subsection{Zero-Short Prompting}
Given the proven effectiveness of zero-shot prompting strategies in various tasks\cite{kojima-2022,reynolds-2021}, including those involving complex reasoning and style-based predictions, we also adopted a zero-shot approach in our experiments, prompting the models with a \textquote{problem}---a sequence of sentences---and a task: predict style changes occurring in pairs of adjacent sentences (0-1, 1-2, etc.).  Following the insights from the literature emphasizing that the carefully engineered prompts are essential for high-quality output \cite{sahoo2025systematicsurveypromptengineering}, we tailored ours in two ways. 

First, after initial experiments with minimal prompt, we tried to employ what can be described as \textquote{strategic guessing}. We began by establishing random baselines\footnote{Random baselines consist of three trivial strategies: never predicting an author change, always predicting a change at every boundary, and randomly selecting a fixed number of change points (e.g. 1, 2, or 3). F1 macro was computed for each strategy, see Table~\ref{tbl_baselines}.} and carefully exploring the data to identify the strategies that could improve the models' odds independently of actual textual content. It turned out that hypothesizing $3$ or $4$ authors (that is, $2$ or $3$ style switches per problem) would cover most cases in the data (see Figure \ref{fig:num_changes} in Appendix \ref{app:exploration}). Therefore, our prompt (see Appendix \ref{app:prompt}) included an explicit recommendation to assume that there are \textquote{approximately 3} authors in the problem. We then trained an XGBoost classifier to predict the number of authors (see Appendix \ref{app:exploration}) and injected its output into the prompt, but it overwhelmingly predicted three authors. As a result, there was no significant performance difference between the static prompt and the dynamic, classifier‐driven prompt.

\begin{table*}[h]
\centering
\caption{Random and constant baselines.}
\label{tbl_baselines}
\begin{tabular}{lc}
\toprule
Baseline & F1 (macro)  \\
\midrule
all changes & $0.1570$ \\
no changes & $0.4426$ \\
$3$ random changes & $0.4946$ \\
$4$ random changes & $0.4982$ \\
\bottomrule
\end{tabular}
\end{table*}

The second axis of our prompting strategy involved providing explicit instructions about the stylistic features expected to be useful for distinguishing between authors\footnote{Our prompt contained the following addition: \textquote{Analyze the writing styles of the input texts, disregarding the differences in topic and content. Base your decision on linguistic features such as: phrasal verbs; modal verbs punctuation; rare words; affixes; quantities; humor; sarcasm; typographical errors; misspellings.}}. For contemporary and well-resourced languages, at least,  such a strategy has been reported to be beneficial \cite{huang_can_2024}\footnote{However, it remains an open question whether this approach retains its effectiveness when applied to historical languages \cite{schmidt_sui_2024}. With reservations, own research suggests the opposite.}, which builds on the conceptual framework outlined in \cite{Grant_2022}. 

\subsection{Measure of \textquote{Correctness}: Hamming Distance}
To assess the factors influencing model predictions within a single problem, one needs a measure to quantify how closely the predicted sequence of author change labels matches the ground truth. Hamming distance is a metric used to measure the difference between two strings of equal length. It is defined as the number of positions at which the corresponding symbols differ. This makes it particularly suitable for evaluating the overall correctness of a model's output in sentence-level style change detection, where each sentence is assigned an author label. For easier interpretation and cross-problem comparison, we use the normalized Hamming distance, obtained by dividing the raw distance by the total number of sentences. A lower value indicates higher prediction accuracy, while a value of zero denotes a perfect match with the ground truth.

\subsection{Measure of Semantic Similarity}

With a clear indicator of overall prediction accuracy for individual problems, we further examine its correlation with various measures of semantic similarity within each problem.  We used \texttt{sentence-transformers/all-MiniLM-L6-v2} and \texttt{styleDistance/styledistance} to vectorize the sentences and compute the semantic similarity and correlations. Specifically, we consider several related metrics:
\begin{enumerate}
    \item Average cosine similarity of sentences within a problem;
    \item average cosine similarity of pairs with an author switch; 
    \item average cosine similarity of adjacent pairs;
    \item mean pairwise cosine distance within a problem.
\end{enumerate}

\subsection{Data}
To ensure the comparability of our results with existing approaches, we rely on the widely recognized evaluation framework of PAN and used their  official \textquote{Multi-Author Writing Style Analysis} datasets from $2024$ and $2025$. The data consists of lists of sentences (such lists are called \textquote{problems}) and arrays of binary labels for each pair of adjacent sentences in the problem. The texts represent continuous discussions in Reddit threads. $2025$ data is divided into sentences, while $2024$ into paragraphs. Both datasets are served at three levels of difficulty: \texttt{easy}, \texttt{medium}, and \texttt{hard}, each \texttt{split} into \texttt{train} ($70$\%), \texttt{validation} ($15$\%), and \texttt{test} ($15$\%) sets. In this study, all exploratory work is done on \texttt{train} set, while \texttt{validation} is only used for prompting LLMs. 

\subsection{Models \& Experiments}
Our experiment comprises three stages. In the first stage, we randomly sample $250$ problems from the \texttt{validation} splits of three datasets and submit them as prompts to state-of-the-art models: \texttt{GPT-4o}, \texttt{Claude 3.7 Sonnet}, \texttt{DeepSeek-R1}, and \texttt{Meta-Llama-3.1-405B-Instruct}. In the second stage, we prompt the entire \texttt{validation} splits of each dataset ($900$ problems) to the best-performing model identified in the first stage. In the third stage, having obtained predictions from that model, we compute the Hamming distance, semantic similarity, and correlations as described above. 

\section{Results}

The results of the first evaluation stage are summarized in Table~\ref{tbl_results}. Among the evaluated models, \texttt{Claude-3.7-Sonnet} consistently outperformed its competitors---\texttt{GPT-4o}, \texttt{Deepseek-R1}, and \texttt{Llama-3.1-405B-Instruct}---across all difficulty levels and despite the distinct characteristics of the three datasets (see Appendix~\ref{app:exploration}). Even in the zero-shot setting, the model's reasoning allowed it to achieve competitive accuracy.
\begin{table*}[ht]
\footnotesize
\centering
\caption{Performance (F1 macro) on $250$ randomly sampled problems (validation split).}
\label{tbl_results}

\begin{tabular}{l@{\quad}c@{\quad}c@{\quad}c}
\toprule
Model & Easy & Medium & Hard \\
\midrule
\texttt{Claude-3.7-Sonnet}     & 0.8638 & 0.8412 & 0.6580 \\
\texttt{DeepSeek-R1}           & 0.6332 & 0.6082 & 0.5263 \\
\texttt{LLaMA-3.1-40B-Instruct} & 0.6817 & 0.6609 & 0.5614 \\
\texttt{GPT-4o}                & 0.5540 & 0.5557 & 0.5105 \\
\bottomrule
\end{tabular}
\end{table*}

Table~\ref{tbl_claude_results} presents detailed performance metrics for the best-performing model, \texttt{Claude-3.7-sonnet}, evaluated on the full validation splits of all three datasets. To contextualize Claude’s performance, we fine-tuned a \texttt{styleDistance/styledistance} transformer combined with an LSTM and MLP (127m) for adjacent sentence classification using combined PAN 2024 and PAN 2025 training data\footnote{For training we used the paragraph-level annotations to generate labeled pairs by splitting paragraphs into adjacent units, treating them as sentence-like segments. This allowed us to augment the data and obtain style-change labels between pairs. }. Remarkably, Claude’s zero-shot prompting performance nearly surpasses that of the fine-tuned transformer on the \texttt{medium} dataset and remains close behind on the other two. These results align with additional observations made on the PAN 2024 paragraph-level style-change dataset, we simply adapted the prompt by replacing the term “sentence” with “paragraph”. There, Claude achieved a score of $0.618$ on the \texttt{hard} dataset, outperforming not only baseline models but also four official submissions \cite{ayele_overview_2024}. Unsurprisingly, Claude’s results on \texttt{easy} and \texttt{medium} datasets were even more impressive, reaching $0.83$.
\begin{table*}[ht]
\centering
\footnotesize
\caption{Performance of \texttt{claude-3.7-sonnet} on full validation data of the three tasks.}
\label{tbl_claude_results}

\begin{tabular}{l@{\quad}c@{\quad}c@{\quad}c}
\toprule
Model & Easy & Medium & Hard \\
\midrule
\texttt{Claude-3.7-Sonnet}                              & 0.8559 & 0.8182 & 0.6612 \\
\texttt{StyleDistance + LSTM + MLP}          & 0.9231 & 0.8276 & 0.7240 \\
\bottomrule
\end{tabular}
\end{table*}

\section{Discussion}
\subsection{When Does Claude Fail?}

Measuring Hamming distance allowed us to examine how prediction accuracy is influenced by various problem-level parameters. Table~\ref{tbl:hamming} presents correlations between Hamming distance and four parameters: the number of authors per problem, the number of actual author changes, the number of predicted changes, and the total number of sentences in the problem.

The number of predicted changes shows a strong positive correlation with Hamming distance across all three datasets. This indicates that the more style changes Claude predicts---whether correctly or not---the more likely it is to deviate from the ground truth. This suggests this tendency of the model to over-segment is a key source of error.

The number of actual style changes is only weakly correlated with Hamming distance, and significantly so only in the easy and medium datasets. This indicates that the intrinsic complexity of the text (in terms of true author switches) only slightly contributes to model confusion, suggesting that the local stylistic context may play a bigger role than the dynamics in the problem in general.

The medium dataset stands out as particularly challenging. The correlation between predicted changes and Hamming distance is strongest here, and the correlation with the number of authors is also positive and significant, unlike in the easy and hard sets. This suggests that the medium dataset, while not explicitly labeled as most difficult, may contain greater internal variance---particularly in the number and distribution of authors---leading to less reliable prediction.

In sum, these correlations seem to bring to light that prediction errors are less a function of the data complexity and more a result of the model’s internal heuristics---especially its tendency to overpredict change points. These findings suggest that there probably remains a room for improvement of the prompt and the entire pipeline, in which the LLM will be a crucial but not only participant.

\begin{table*}[h]
\centering
\caption{Correlation between Hamming distance of predicted vs. actual style switches and problem-level parameters.}
\label{tbl:hamming}
\begin{tabular}{llrrr}
\toprule
Dataset & Feature & Spearman & Pearson & Kendall \\
\midrule
\multirow{4}{*}{easy} 
& num\_authors        & $-0.056$  & $-0.078$ & $-0.045$  \\
& num\_changes        & $-0.092$ & $\mathbf{-0.104}$ & $-0.073$  \\
& num\_changes\_pred  & $\mathbf{0.400}$ & $\mathbf{0.387}$ & $\mathbf{0.312}$  \\
& num\_sentences      & $\mathbf{-0.098}$  & $\mathbf{-0.122}$ & $-0.099$ \\
\midrule
\multirow{4}{*}{hard}
& num\_authors        & $0.048$  & $0.036$  & $0.037$  \\
& num\_changes        & $0.056$ & $0.051$ & $0.043$  \\
& num\_changes\_pred  & $\mathbf{0.337}$ & $\mathbf{0.303}$ & $\mathbf{0.257}$  \\
& num\_sentences      & $\mathbf{-0.260}$ & $\mathbf{-0.277}$ & $\mathbf{-0.197}$  \\
\midrule
\multirow{4}{*}{medium}
& num\_authors        & $\mathbf{0.168}$  & $\mathbf{0.141}$ & $\mathbf{0.134}$  \\
& num\_changes        & $\mathbf{0.178}$ & $\mathbf{0.176}$ & $\mathbf{0.134}$ \\
& num\_changes\_pred  & $\mathbf{0.455}$  & $\mathbf{0.400}$ &  $\mathbf{0.346}$ \\
& num\_sentences      & $\mathbf{0.140}$ & $\mathbf{0.132}$ &  $\mathbf{0.083}$  \\
\bottomrule
\end{tabular}
\end{table*}

\subsection{Correlation between style change and semantics}

A comparison between correlations of semantic similarity with predicted and true authorship changes reveals a striking pattern. Importantly, note that a predicted value of 1 indicates a change of author, which is expected to correspond to low cosine similarity between adjacent sentences; conversely, a value of 0 (no change) should align with high similarity. Therefore, a negative correlation implies that model predictions are consistent with this expectation --- the lower the similarity, the more likely the model predicts a change.

Across all models, we observe a moderate to strong correlation on the easy split, suggesting that stylistic shifts are often accompanied by semantic divergence in simpler cases. However, this correlation becomes much weaker or disappears entirely on the hard split, indicating that in more complex texts, stylistic changes do not always show as measurable differences in sentence embeddings, see Table \ref{tbl_cosine}.

Moreover, in the ground truth, the interesting trend is observed in harder cases: true authorship changes often occur between semantically similar sentences, possibly due to topic continuity or deliberate stylistic imitation.


\begin{table}[h]
\centering
\footnotesize
\caption{Spearman, Pearson, and Kendall correlation between sentence similarity and a switch prediction.}
\label{tbl_cosine}
\begin{tabular}{lccc}
\toprule
Dataset & Spearman & Pearson & Kendall \\
\midrule
easy    & $\mathbf{-0.239}$ & $\mathbf{-0.205}$ & $\mathbf{-0.195}$ \\
medium  & $\mathbf{-0.160}$ & $\mathbf{-0.148}$ & $\mathbf{-0.131}$ \\
hard    & $\mathbf{-0.117}$ & $\mathbf{-0.102}$ & $\mathbf{-0.096}$ \\
\bottomrule
\end{tabular}
\end{table}

\subsection{Generalizability of Suggested Approach}

In our prompting strategy, we explicitly instructed the model to assume the presence of approximately three authors per problem. This decision was informed by a statistical analysis of the PAN datasets, which revealed that the majority of documents contain between two and four authors (see Appendix \ref{app:exploration}, Figure \ref{fig:num_changes}). However, we acknowledge that this heuristic is highly dependent on the specific distribution of the PAN data. When applying the same approach to corpora from other domains—such as literary texts, social media outside Reddit, or historical documents—it would be necessary to conduct a similar exploratory analysis to determine an appropriate prior. While effective in our case, such assumptions should be treated as dataset-specific and not be blindly generalized across tasks.

\subsection{LLMs: A Baseline for PAN?}
An important consideration in interpreting our results is the source of the data. The documents used in this task originate from Reddit, a platform whose content has probably been extensively represented in the training data of the LLMs. This raises the possibility that the models, despite being used as black boxes, may benefit from prior exposure to the domain and its stylistic conventions. However, rather than undermining our findings, this observation reinforces a central claim of our study: LLMs perform surprisingly well on the multi-author style change detection task in a zero-shot setting, relying solely on prompt-based guidance and without any task-specific fine-tuning. On this basis, we argue that LLMs should be regarded as a strong and competitive baseline for this task.

This observation, in turn, raises broader questions about task design. If the goal is to rigorously assess the generalization capabilities of LLMs, particularly in unfamiliar stylistic or genre contexts, future benchmarks may need to draw on curated or out-of-domain sources. Historical corpora, literary texts, or non-mainstream digital genres could provide more robust tests of the models' true stylistic sensitivity, independent of memorization or domain familiarity.

\subsection{Hallucinations}

LLMs are increasingly optimized to minimize hallucinations. Prominent models such as GPT and Claude incorporate internal safety mechanisms aiming at ethical alignment and factual reliability. While these safeguards are not always thoroughly documented in technical papers, they can be broadly understood as a set of training procedures and architectural decisions geared towards reducing harmful, irrelevant, or incorrect outputs.

A central component of these efforts is reinforcement learning from human feedback (RLHF) \cite{lambert2022illustrating,ouyang2022traininglanguagemodelsfollow}, which plays a critical role in aligning model behavior with human preferences, including truthfulness and ethical reasoning. This alignment process has been shown to reduce hallucinations and improve the model's ability to reference factual content with greater accuracy.

Recent research has begun to explore the extent to which LLMs can attribute their outputs to external sources—a capability closely related to hallucination minimization. For example, Gao et al.~\cite{gao_enabling_2023} introduce ALCE, a benchmark for evaluating citation quality in LLM-generated answers. ALCE combines metrics for fluency, factuality, and attribution across QA datasets. Similarly, AttrScore \cite{yue-etal-2023-automatic} offers a framework for evaluating whether model-generated claims are supported by cited references, using classification labels such as attributable, extrapolatory, and contradictory. This approach integrates both prompting and fine-tuning strategies, and is validated on both synthetic and real-world Bing search outputs.

Further, a comprehensive survey by Li et al.\cite{li_survey_2023} reviews existing techniques for tracing LLM outputs back to source materials, presenting an overview of available datasets, evaluation metrics, error typologies, and outstanding challenges in grounding generated content in verifiable evidence. Complementary to this, Guo et al.\cite{guo-etal-2022-survey} propose a unified framework for automated fact-checking, encompassing claim detection, evidence retrieval, verdict prediction, and justification generation, along with a systematic review of associated resources.

It is reasonable to hypothesize that a model's attribution capability may correlate with its hallucination rate, with stronger grounding mechanisms contributing to reduced factual errors. However, validating this hypothesis is difficult due to several factors: the proprietary nature of leading LLMs, limited transparency in model architecture and training data, and the lack of standardized, robust benchmarks for hallucination detection in state-of-the-art systems. This remains a critical limitation for future research, especially in high-stakes applications requiring verifiability and trustworthiness.

\bibliographystyle{splncs04}
\bibliography{references, custom}

\begin{thebibliography}{10}
\providecommand{\url}[1]{\texttt{#1}}
\providecommand{\urlprefix}{URL }
\providecommand{\doi}[1]{https://doi.org/#1}

\bibitem{ayele_overview_2024}
Ayele, A.A., Babakov, N., Bevendorff, J., Casals, X.B., Chulvi, B., Dementieva, D., Elnagar, A., Freitag, D., Fröbe, M., Korenčić, D., Mayerl, M., Moskovskiy, D., Mukherjee, A., Panchenko, A., Potthast, M., Rangel, F., Rizwan, N., Rosso, P., Schneider, F., Smirnova, A., Stamatatos, E., Stakovskii, E., Stein, B., Taulé, M., Ustalov, D., Wang, X., Wiegmann, M., Yimam, S.M., Zangerle, E.: Overview of {PAN} 2024: {Multi}-{Author} {Writing} {Style} {Analysis}, {Multilingual} {Text} {Detoxification}, {Oppositional} {Thinking} {Analysis}, and {Generative} {AI} {Authorship} {Verification}. In: Goeuriot, L., Mulhem, P., Quénot, G., Schwab, D., Nunzio, G.M.D., Soulier, L., Galuscakova, P., Herrera, A.G.S., Faggioli, G., Ferro, N. (eds.) Experimental {IR} {Meets} {Multilinguality}, {Multimodality}, and {Interaction}. 15th {International} {Conference} of the {CLEF} {Association} ({CLEF} 2024). Lecture {Notes} in {Computer} {Science}, Springer, Berlin Heidelberg New York (Sep 2024)

\bibitem{bagnall_authorship_2016}
Bagnall, D.: Authorship clustering using multi-headed recurrent neural networks  (2016). \doi{10.48550/ARXIV.1608.04485}, \url{https://arxiv.org/abs/1608.04485}, publisher: arXiv Version Number: 1

\bibitem{gao_enabling_2023}
Gao, T., Yen, H., Yu, J., Chen, D.: Enabling {Large} {Language} {Models} to {Generate} {Text} with {Citations}. In: Bouamor, H., Pino, J., Bali, K. (eds.) Proceedings of the 2023 {Conference} on {Empirical} {Methods} in {Natural} {Language} {Processing}. pp. 6465--6488. Association for Computational Linguistics, Singapore (Dec 2023). \doi{10.18653/v1/2023.emnlp-main.398}, \url{https://aclanthology.org/2023.emnlp-main.398/}

\bibitem{graner_unorthodox_2022}
Graner, L., Ranly, P.: An {Unorthodox} {Approach} for {Style} {Change} {Detection}. In: {CLEF} ({Working} {Notes}). pp. 2455--2466 (2022)

\bibitem{Grant_2022}
Grant, T.: The Idea of Progress in Forensic Authorship Analysis. Elements in Forensic Linguistics, Cambridge University Press (2022)

\bibitem{guo-etal-2022-survey}
Guo, Z., Schlichtkrull, M., Vlachos, A.: A survey on automated fact-checking. Transactions of the Association for Computational Linguistics  \textbf{10},  178--206 (2022). \doi{10.1162/tacl_a_00454}, \url{https://aclanthology.org/2022.tacl-1.11/}

\bibitem{gomez-adorno_author_2017}
Gómez-Adorno, H., Aleman, Y., Ayala, D.V., Sanchez-Perez, M.A., Pinto, D., Sidorov, G.: Author {Clustering} using {Hierarchical} {Clustering} {Analysis}. In: {CLEF} ({Working} notes) (2017)

\bibitem{hashemi_enhancing_2023}
Hashemi, A., Shi, W.: Enhancing {Writing} {Style} {Change} {Detection} using {Transformer}-based {Models} and {Data} {Augmentation}. In: {CLEF} ({Working} {Notes}). pp. 2613--2621 (2023)

\bibitem{he2021deberta}
He, P., Liu, X., Gao, J., Chen, W.: Deberta: Decoding-enhanced bert with disentangled attention. In: International Conference on Learning Representations (2021), \url{https://openreview.net/forum?id=XPZIaotutsD}

\bibitem{huang_can_2024}
Huang, B., Chen, C., Shu, K.: Can {Large} {Language} {Models} {Identify} {Authorship}? (Oct 2024). \doi{10.48550/arXiv.2403.08213}, \url{http://arxiv.org/abs/2403.08213}, arXiv:2403.08213 [cs]

\bibitem{huang_authorship_2025}
Huang, B., Chen, C., Shu, K.: Authorship {Attribution} in the {Era} of {LLMs}: {Problems}, {Methodologies}, and {Challenges} (2025), \url{https://arxiv.org/abs/2408.08946}, \_eprint: 2408.08946

\bibitem{hung-etal-2023-wrote}
Hung, C.Y., Hu, Z., Hu, Y., Lee, R.: Who wrote it and why? prompting large-language models for authorship verification. In: Bouamor, H., Pino, J., Bali, K. (eds.) Findings of the Association for Computational Linguistics: EMNLP 2023. pp. 14078--14084. Association for Computational Linguistics (2023). \doi{10.18653/v1/2023.findings-emnlp.937}, \url{https://aclanthology.org/2023.findings-emnlp.937/}

\bibitem{kocher_author_2017}
Kocher, M., Savoy, J.: Author clustering with an adaptive threshold. In: Experimental {IR} {Meets} {Multilinguality}, {Multimodality}, and {Interaction}: 8th {International} {Conference} of the {CLEF} {Association}, {CLEF} 2017, {Dublin}, {Ireland}, {September} 11–14, 2017, {Proceedings} 8. pp. 186--198. Springer (2017)

\bibitem{kojima-2022}
Kojima, T., Gu, S.S., Reid, M., Matsuo, Y., Iwasawa, Y.: Large language models are zero-shot reasoners. In: Proceedings of the 36th International Conference on Neural Information Processing Systems. NIPS '22, Curran Associates Inc., Red Hook, NY, USA (2022)

\bibitem{kushnareva-etal-2021-artificial}
Kushnareva, L., Cherniavskii, D., Mikhailov, V., Artemova, E., Barannikov, S., Bernstein, A., Piontkovskaya, I., Piontkovski, D., Burnaev, E.: Artificial text detection via examining the topology of attention maps. In: Moens, M.F., Huang, X., Specia, L., Yih, S.W.t. (eds.) Proceedings of the 2021 Conference on Empirical Methods in Natural Language Processing. pp. 635--649. Association for Computational Linguistics, Online and Punta Cana, Dominican Republic (2021). \doi{10.18653/v1/2021.emnlp-main.50}, \url{https://aclanthology.org/2021.emnlp-main.50/}

\bibitem{lambert2022illustrating}
Lambert, N., Castricato, L., von Werra, L., Havrilla, A.: Illustrating reinforcement learning from human feedback (rlhf). Hugging Face Blog  (2022), https://huggingface.co/blog/rlhf

\bibitem{li_survey_2023}
Li, D., Sun, Z., Hu, X., Liu, Z., Chen, Z., Hu, B., Wu, A., Zhang, M.: A {Survey} of {Large} {Language} {Models} {Attribution} (2023), \url{https://arxiv.org/abs/2311.03731}, \_eprint: 2311.03731

\bibitem{lin_team_2024}
Lin, T., Wu, Y., Lee, L.: Team {NYCU}-{NLP} at {PAN} 2024: integrating transformers with similarity adjustments for multi-author writing style analysis. Working Notes of CLEF  (2024)

\bibitem{lo-etal-2021-transformer-pre}
Lo, K., Jin, Y., Tan, W., Liu, M., Du, L., Buntine, W.: Transformer over pre-trained transformer for neural text segmentation with enhanced topic coherence. In: Moens, M.F., Huang, X., Specia, L., Yih, S.W.t. (eds.) Findings of the Association for Computational Linguistics: EMNLP 2021. pp. 3334--3340. Association for Computational Linguistics (2021). \doi{10.18653/v1/2021.findings-emnlp.283}, \url{https://aclanthology.org/2021.findings-emnlp.283/}

\bibitem{ouyang2022traininglanguagemodelsfollow}
Ouyang, L., Wu, J., Jiang, X., Almeida, D., Wainwright, C.L., Mishkin, P., Zhang, C., Agarwal, S., Slama, K., Ray, A., Schulman, J., Hilton, J., Kelton, F., Miller, L., Simens, M., Askell, A., Welinder, P., Christiano, P., Leike, J., Lowe, R.: Training language models to follow instructions with human feedback. arXiv  (2022), \url{https://arxiv.org/abs/2203.02155}

\bibitem{patel_learning_2023}
Patel, A., Rao, D., Kothary, A., McKeown, K., Callison-Burch, C.: Learning {Interpretable} {Style} {Embeddings} via {Prompting} {LLMs} (Oct 2023), \url{http://arxiv.org/abs/2305.12696}, arXiv:2305.12696 [cs]

\bibitem{potthast_overview_2017}
Potthast, M., Rangel, F., Tschuggnall, M., Stamatatos, E., Rosso, P., Stein, B.: Overview of {PAN} 2017: {Author} {Identification}, {Author} {Profiling}, and {Author} {Obfuscation}. In: Jones, G.J.F., Lawless, S., Gonzalo, J., Kelly, L., Goeuriot, L., Mandl, T., Cappellato, L., Ferro, N. (eds.) Experimental {IR} {Meets} {Multilinguality}, {Multimodality}, and {Interaction}. 8th {International} {Conference} of the {CLEF} {Initiative} ({CLEF} 2017). Lecture {Notes} in {Computer} {Science}, vol. 10456, pp. 275--290. Springer, Berlin Heidelberg New York (Sep 2017)

\bibitem{ramnath_cave_2024}
Ramnath, S., Pandey, K., Boschee, E., Ren, X.: {CAVE}: {Controllable} {Authorship} {Verification} {Explanations}. arXiv preprint arXiv:2406.16672  (2024)

\bibitem{reynolds-2021}
Reynolds, L., McDonell, K.: Prompt programming for large language models: Beyond the few-shot paradigm. In: Extended Abstracts of the 2021 CHI Conference on Human Factors in Computing Systems. CHI EA '21, Association for Computing Machinery, New York, NY, USA (2021). \doi{10.1145/3411763.3451760}, \url{https://doi.org/10.1145/3411763.3451760}

\bibitem{sahoo2025systematicsurveypromptengineering}
Sahoo, P., Singh, A.K., Saha, S., Jain, V., Mondal, S., Chadha, A.: A systematic survey of prompt engineering in large language models: Techniques and applications (2025), \url{https://arxiv.org/abs/2402.07927}

\bibitem{schmidt_sui_2024}
Schmidt, G., Gorovaia, S., Yamshchikov, I.: Sui {Generis}: {Large} {Language} {Models} for {Authorship} {Attribution} and {Verification} in {Latin}. Miami, FL (Nov 2024)

\bibitem{stamatatos_clustering_2016}
Stamatatos, E., Tschnuggnall, M., Verhoeven, B., Daelemans, W., Specht, G., Stein, B., Potthast, M.: Clustering by authorship within and across documents. In: Working {Notes} {Papers} of the {CLEF} 2016 {Evaluation} {Labs}. {CEUR} {Workshop} {Proceedings}/{Balog}, {Krisztian} [edit.]; et al. pp. 691--715 (2016)

\bibitem{tolstykh2024gigacheckdetectingllmgeneratedcontent}
Tolstykh, I., Tsybina, A., Yakubson, S., Gordeev, A., Dokholyan, V., Kuprashevich, M.: Gigacheck: Detecting llm-generated content (2024), \url{https://arxiv.org/abs/2410.23728}

\bibitem{uchendu-2024}
Uchendu, A., Le, T., Lee, D.: TopFormer: Topology-Aware Authorship Attribution of Deepfake Texts with Diverse Writing Styles (2024). \doi{10.3233/FAIA240647}

\bibitem{wang-etal-2023-seqxgpt}
Wang, P., Li, L., Ren, K., Jiang, B., Zhang, D., Qiu, X.: {S}eq{XGPT}: Sentence-level {AI}-generated text detection. In: Bouamor, H., Pino, J., Bali, K. (eds.) Proceedings of the 2023 Conference on Empirical Methods in Natural Language Processing. pp. 1144--1156. Association for Computational Linguistics, Singapore (2023), \url{https://aclanthology.org/2023.emnlp-main.73/}

\bibitem{wu3-2024}
Wu, Q., Kong, L., Ye, Z.: {Team bingezzzleep at PAN: A Writing Style Change Analysis Model Based on RoBERTa Encoding and Contrastive Learning for Multi-Author Writing Style Analysis}. In: Faggioli, G., Ferro, N., Galu{\v{s}}{\v{c}}{\'a}kov{\'a}, P., Herrera, A.G.S. (eds.) Working Notes Papers of the CLEF 2024 Evaluation Labs. pp. 2963--2968. CEUR-WS.org (Sep 2024), \url{http://ceur-ws.org/Vol-3740/paper-288.pdf}

\bibitem{ye_supervised_2023}
Ye, Z., Zhong, C., Qi, H., Han, Y.: Supervised {Contrastive} {Learning} for {Multi}-{Author} {Writing} {Style} {Analysis}. In: {CLEF} ({Working} {Notes}). pp. 2817--2822 (2023)

\bibitem{yue-etal-2023-automatic}
Yue, X., Wang, B., Chen, Z., Zhang, K., Su, Y., Sun, H.: Automatic evaluation of attribution by large language models. In: Bouamor, H., Pino, J., Bali, K. (eds.) Findings of the Association for Computational Linguistics: EMNLP 2023. pp. 4615--4635. Association for Computational Linguistics (2023). \doi{10.18653/v1/2023.findings-emnlp.307}, \url{https://aclanthology.org/2023.findings-emnlp.307/}

\bibitem{zangerle_overview_2023}
Zangerle, E., Mayerl, M., Potthast, M., Stein, B.: Overview of the {Multi}-{Author} {Writing} {Style} {Analysis} {Task} at {PAN} 2023. In: Aliannejadi, M., Faggioli, G., Ferro, N., Vlachos, M. (eds.) Working {Notes} of the {Conference} and {Labs} of the {Evaluation} {Forum} ({CLEF} 2023). {CEUR} {Workshop} {Proceedings}, vol.~3497, pp. 2513--2522 (Sep 2023), \url{https://ceur-ws.org/Vol-3497/paper-201.pdf}

\bibitem{zheng-2024-detecting-ai}
Zeng, Z., Liu, S., Sha, L., Li, Z., Yang, K., Liu, S., Ga\v{s}evi\'{c}, D., Chen, G.: Detecting ai-generated sentences in human-ai collaborative hybrid texts: challenges, strategies, and insights. In: Proceedings of the Thirty-Third International Joint Conference on Artificial Intelligence. IJCAI '24 (2024). \doi{10.24963/ijcai.2024/835}, \url{https://doi.org/10.24963/ijcai.2024/835}

\bibitem{zeng2024detectingaigeneratedsentenceshumanai}
Zeng, Z., Liu, S., Sha, L., Li, Z., Yang, K., Liu, S., Gašević, D., Chen, G.: Detecting ai-generated sentences in human-ai collaborative hybrid texts: Challenges, strategies, and insights (2024), \url{https://arxiv.org/abs/2403.03506}

\bibitem{zhang_style_2022}
Zhang, Z., Han, Z., Kong, L.: Style {Change} {Detection} based on {Prompt}. In: {CLEF} ({Working} {Notes}). pp. 2753--2756 (2022)

\bibitem{zhuang-etal-2021-robustly}
Zhuang, L., Wayne, L., Ya, S., Jun, Z.: A robustly optimized {BERT} pre-training approach with post-training. In: Li, S., Sun, M., Liu, Y., Wu, H., Liu, K., Che, W., He, S., Rao, G. (eds.) Proceedings of the 20th Chinese National Conference on Computational Linguistics. pp. 1218--1227. Chinese Information Processing Society of China (2021), \url{https://aclanthology.org/2021.ccl-1.108/}

\end{thebibliography}

\appendix
\section{Prompt Template Used for LLM Inference}
\label{app:prompt}
\begin{lstlisting}[language=Python, caption={LLM Prompt Template}]
PROMPT_TEMPLATE = """You are an expert in authorship attribution. You notice changes in writing style, topic and especially in tone and sentiment and use this information to complete your task.
Your task is to analyze a sequence of sentences and determine a binary sequence indicating where the author changes at sentence boundaries (1 = change, 0 = no change).
The input is a json formatted list of sentences, which can include quotes, which are escaped with backslashes.
There is always at least one change present.
Return your response in the following JSON format: 
{ "changes": [<0s and 1s indicating sentence-boundary changes>] }

Example 1:
Input: ["Sentence one.", "Sentence two.", "Sentence three.", "Sentence four."]
Output: { "changes": [0, 1, 0] }

Example 2:
Input: ["This is written by author A.", "Another sentence by author A.", "Alright... now, a new author starts.", "Yet another author change here."]
Output: { "changes": [0, 1, 1] }

Now analyze the following text:"""

DYNAMIC_PROMPT_PART = (
    "There should be exactly {} sentences resulting in a changes array of lenght {}. 
    
    Keep in mind, that the sentences originate from reddit, so consecutive sentences, that agree with each other do not always have the same author, if the tone or style changes. Still, consider that they may follow each other but always assume around 3 changes. Analyze the writing styles of the input texts, disregarding the differences in topic and content. Base your decision on linguistic features such as: phrasal verbs; modal verbs punctuation; rare words; affixes; quantities; humor; sarcasm; typographical errors; misspellings."
)
\end{lstlisting}

\section{Data Exploration} \label{app:exploration}

We analyzed the distribution of authors per document, revealing a balanced composition primarily dominated by documents featuring two to four authors, thus typically containing one to three style changes. This finding informed our prompt design, as we explicitly indicated the expected number of style switches to guide model predictions.

\begin{figure}[ht]
    \centering
    \includegraphics[width=\textwidth]{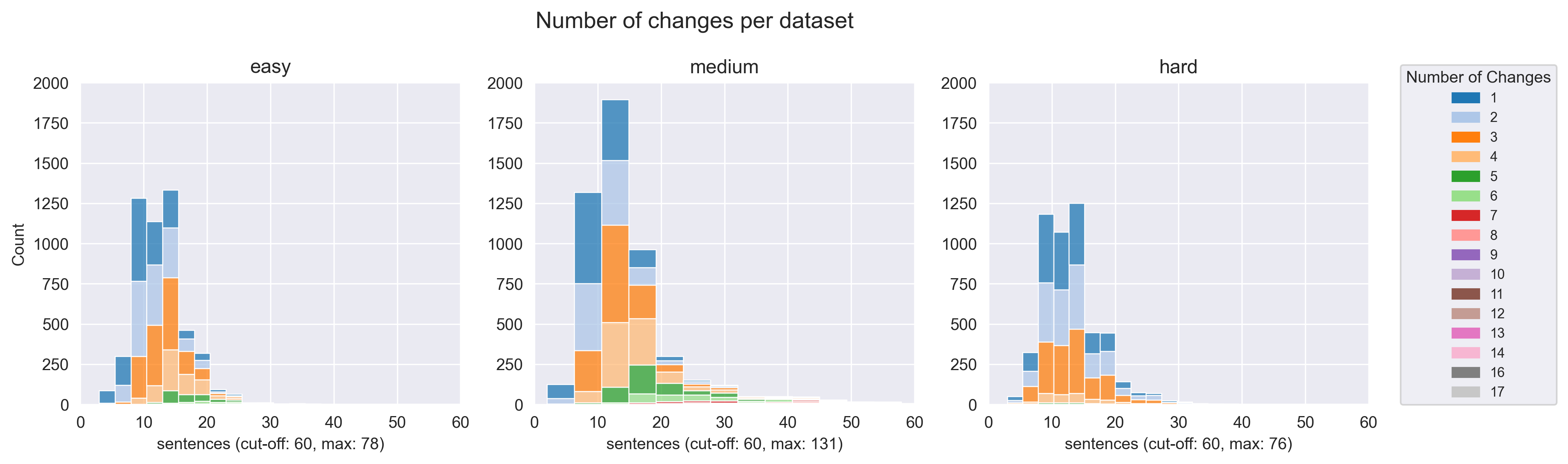}
    \caption{Number of changes per dataset.}
    \label{fig:num_changes}
\end{figure}

Notably, categorical labels representing the presence of either \textquote{1–2} or \textquote{3–4} authors per document could be predicted with reasonable accuracy using only superficial features. In our experiment, a general description of cosine similarity distributions between adjacent sentences within a problem together with a fixed-size binned representation of this distribution and basic stylistic features extracted via the \texttt{textstat} package allowed us to obtain reasonable results on \texttt{easy} and \texttt{medium} datasets.

Table \ref{tbl_num_changes} indicates significant potential for developing a system that provides models with highly reliable and precise information regarding the expected number of authors.

\begin{table*}[ht]
\centering
\caption{Performance of \texttt{XGBoostClassifier} on the number-of-changes prediction tasks (as categorical variable). Mean over $100$ stratified shuffled splits (20\% of the data).} \label{tbl_num_changes}
\begin{tabular}{llrrr}
\toprule
\textbf{Dataset} & \textbf{Metric} & \textbf{1–2} & \textbf{3–4} & \textbf{5+} \\
\midrule
\multirow{4}{*}{Easy}
  & Precision & 0.832 & 0.680 & 0.543 \\
  & Recall    & 0.818 & 0.716 & 0.437 \\
  & F1-score  & 0.825 & 0.698 & 0.484 \\
  & Support   & 845   & 598   & 87 \\

\midrule
\multirow{4}{*}{Medium}
  & Precision & 0.738 & 0.581 & 0.677 \\
  & Recall    & 0.679 & 0.682 & 0.526 \\
  & F1-score  & 0.707 & 0.628 & 0.592 \\
  & Support   & 660   & 623   & 247 \\
\midrule
\multirow{4}{*}{Hard}
  & Precision & 0.742 & 0.624 & 0.143 \\
  & Recall    & 0.861 & 0.446 & 0.071 \\
  & F1-score  & 0.797 & 0.520 & 0.095 \\
  & Support   & 985   & 531   & 14 \\
\bottomrule

\end{tabular}
\end{table*}

Additionally, we computed pairwise cosine similarities between adjacent sentences, stratified by dataset difficulty levels. Mean cosine similarity generally decreased as the number of style changes increased, although this trend was less pronounced in the hardest dataset.
 
\end{document}